\documentclass[letterpaper]{article} 
\usepackage[preprint]{aaai2027}  
\usepackage[hyphens]{url}  
\usepackage{graphicx} 
\usepackage{natbib} 
\usepackage{caption} 
\usepackage{amsmath}
\usepackage{amssymb}
\usepackage{booktabs}
\usepackage{multirow}
\usepackage{array}
\usepackage{pifont}
\usepackage{algorithm}
\usepackage{algorithmic}

\newcommand{\COCOImages}{500}
\newcommand{\COCOClasses}{80}
\newcommand{\COCOPresentMean}{2.894}
\newcommand{\COCOPresentMedian}{2}
\newcommand{\COCOObjectMean}{7.018}
\newcommand{\COCOObjectMedian}{4}
\newcommand{\COCOZeroRowRate}{96.38}
\newcommand{\COCOMaxClassCount}{13}

\title{UniCounting: Instance-Aware Proposal Consolidation for Image-Query-Free Multi-Category Counting}
\author{Jinshi Liu\textsuperscript{1}, Pan Liu\textsuperscript{2}, Lei He\textsuperscript{3}, Weichao Luo\textsuperscript{4,*}, Rui Qian\textsuperscript{5,*}}
\affiliations{
    \textsuperscript{1}Shenzhen University\quad
    \textsuperscript{2}The Hong Kong University of Science and Technology (Guangzhou)\\
    \textsuperscript{3}Hunan University of Science and Technology\quad
    \textsuperscript{4}Peng Cheng Laboratory\quad
    \textsuperscript{5}Fudan University\\
    \textsuperscript{1}jsl@szu.edu.cn,\quad
    \textsuperscript{2}liup28292@gmail.com,\quad
    \textsuperscript{3}helei\_xb@hnust.edu.cn\\
    \textsuperscript{4}luowch@pcl.ac.cn,\quad
    \textsuperscript{5}qiianruii@gmail.com\\
    \textsuperscript{*}Co-corresponding authors.
}

\begin{document}
\maketitle

\begin{abstract}
Visual counting is commonly formulated as counting a single specified target, with a model receiving an image-specific exemplar, text query, or target category and returning a single count.
We instead study fixed-vocabulary
image-query-free multi-category counting. A global vocabulary is fixed
for each run, and, given only an RGB image, the model predicts a
complete category--count vector without being told which categories
appear.
We present UniCounting, which casts counting as instance-aware
structural inference over an over-complete proposal set. Generic
segmenters produce duplicate masks, partial views, and proposals from
neighboring instances; semantic scores can name them but cannot
determine which denote the same object. Frozen SAM~2.1 generates masks,
while frozen DINOv2 and OpenCLIP provide relation and category features.
A 3,267-parameter category-shared relation head predicts same-instance
affinities from instance-mask-derived supervision. Sparse graph
construction, representative selection, labeling, and background-margin
admission then convert each admitted component into one count with
replayable group evidence. Only the relation head is trained, without
count or density-map targets.
On COCO clean500, UniCounting obtains lower point-estimate vector
$\ell_1$ error and absent-class false mass than calibrated OWLv2-All80,
with comparable micro presence F1. Under a matched decoder, the learned
relation reduces both errors relative to mask containment, mask IoU,
CLIP, and DINO, while revealing a fragmentation--merge trade-off. We
also report transfer diagnostics on OmniCount-sub, FSC-147, and CARPK.
\end{abstract}

\begin{figure}[t]
\centering
\includegraphics[width=0.47\textwidth]{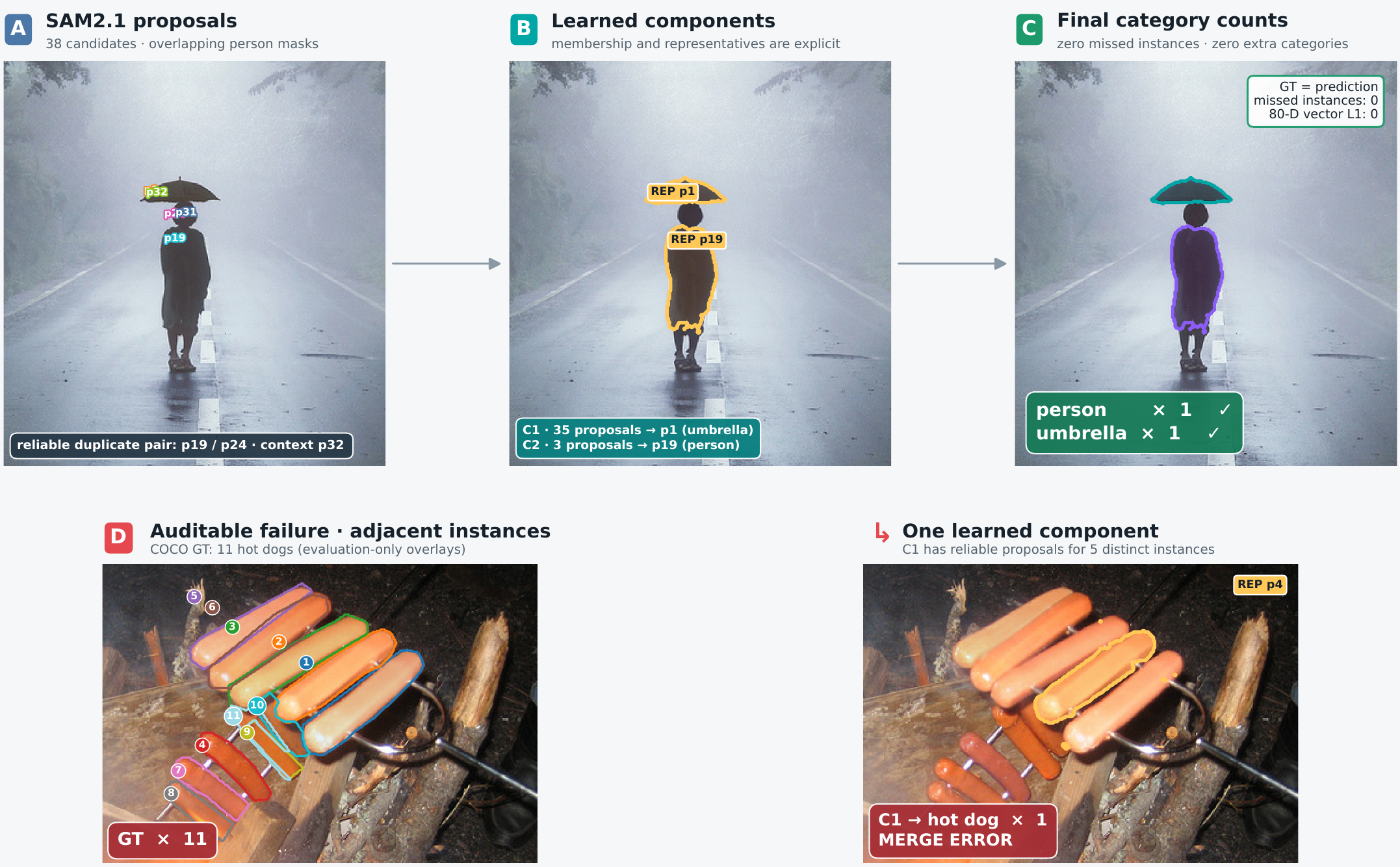}
\caption{Qualitative counting examples. In the upper example, colored boxes mark representative proposals for the displayed groups, and arrows connect each representative to its fixed-vocabulary category and predicted count. In the lower example, one predicted component merges proposals supporting at least five distinct ground-truth hot-dog instances. The remaining discrepancy between the ground-truth count of 11 and the single predicted component may also include upstream proposal misses.}
\label{fig:fig1}
\end{figure}

\section{Introduction}
\label{sec:intro}

Visual counting is typically formulated as a conditional, single-category task. A model receives an exemplar or a category query that specifies what to count and returns a scalar count or density map for that target \citep{ranjan2021famnet,liu2022countr,jiang2023clipcount,amini2024countgd}. This interface is effective when the target is known in advance, but it does not directly answer a scene-level question: which object categories are present, and how many instances of each occur? We study image-query-free multi-category counting, where a global vocabulary is fixed once for each deployment or evaluation run, but no image-dependent category selection is provided. Thus, image-query-free does not mean vocabulary-free; rather, the model receives no per-image target category, exemplar, or restricted class list. Given only an RGB image, it returns a complete category--count vector together with the proposal groups supporting its nonzero entries.

Several lines of work provide parts of this capability. Category-labeled detectors can be converted into count vectors by summing detections, supervised multi-class counters predict fixed category channels, and reference-less counters can organize foreground objects into image-specific groups \citep{minderer2023owlv2,xu2021multiclass,hobley2024abc123,spanakis2026occam}. What remains underexplored is the instance-consolidation problem that appears when counting is built from generic segmentation proposals without an image-specific target. Automatic segmenters produce over-complete candidate sets containing duplicate masks for one object, partial views of the same object, and visually similar proposals from neighboring objects. Semantic scores can assign category names to proposals, but they do not determine whether two proposals denote the same physical instance. Reliable counting therefore requires resolving proposal identity before selecting one representative and one label for each surviving group.

We introduce \textbf{UniCounting}, an instance-aware proposal-consolidation framework built on frozen foundation-model features. SAM 2.1 generates candidate masks, while DINOv2 and OpenCLIP provide relation and category features \citep{ravi2024sam2,oquab2024dinov2,radford2021clip,cherti2023openclip}. A 3,267-parameter category-shared relation head learns pairwise same-instance affinity from instance-mask-derived supervision. UniCounting converts these affinities into a sparse graph, forms connected components, selects a maximum-category-margin representative for each component, assigns the representative's fixed-vocabulary label, and applies background-margin admission using a validation-calibrated threshold. The distinction is not the absence of a vocabulary, but whether image-dependent category selection is supplied. Each admitted component contributes one count. The resulting vector is therefore accompanied by an explicit trail from proposals and relation edges to components, representatives, labels, and admission decisions. Only the relation head is trained; the category head and the three foundation-model components remain frozen, and training uses no count or density-map targets. Figure~\ref{fig:fig1} illustrates this proposal-to-group-to-count trace.

We evaluate UniCounting at two complementary levels. At the system level, complete-vector experiments measure category-wise counting, absent-category errors, and presence prediction, including a calibrated category-labeled detector as a direct reference. At the mechanism level, a matched-decoder study holds the candidate pool and downstream decoder fixed while replacing only the edge score; group-partition diagnostics and component-wise ablations then localize the effects of relation scoring, representative selection, labeling, and admission. Our primary evidence comes from the complete-vector COCO evaluation and the matched-decoder analysis, while the remaining held-out and single-coordinate datasets serve as supporting diagnostics.

Our contributions are twofold:
\begin{itemize}
\item We cast fixed-vocabulary image-query-free multi-category counting as an instance-aware proposal-consolidation problem and introduce UniCounting, a foundation-model pipeline with a 3,267-parameter trainable relation head to convert over-complete frozen proposals into complete count vectors with replayable group evidence.
\item We provide a controlled empirical framework that separates system-level counting performance from proposal-consolidation quality through calibrated complete-vector references, matched-decoder edge comparisons, group-partition diagnostics, and decoder ablations. Using this framework, we show that learned instance relations reduce vector $\ell_1$ error and absent-class false mass relative to the tested edge-score alternatives, while exposing a trade-off between proposal fragmentation and instance merging.
\end{itemize}

\section{Related Work}
\label{sec:related}

\paragraph{Queried counting.}
Few-shot counters identify one target from exemplar boxes \citep{ranjan2021famnet,shi2022bmnet,liu2022countr,dukic2023loca}. Text-conditioned methods instead use a category phrase \citep{jiang2023clipcount,amini2023countx,kang2024vlcounter,amini2024countgd}. PseCo combines such conditioning with proposals and SAM masks \citep{huang2024pseco}. Recent text-guided zero-shot counters further improve cross-modal grounding and exemplar construction \citep{qian2025t2icount,liu2025countse}. In each case, target selection remains external to the counter.

\paragraph{Closed-set and discovered categories.}
Supervised multi-class counting predicts one density channel per trained class \citep{xu2021multiclass,gao2024nwpumoc}, while SIMCO clusters emergent foreground types \citep{godi2020simco}. These establish fixed-channel and category-discovery precedents, so fixed-vocabulary output alone is not our novelty claim.

\paragraph{Detection, grounding, and count vectors.}
Fixed-category detectors provide class-labeled instances, so summing retained detections by class is a direct complete-vector baseline. Grounding DINO takes category names or referring expressions for open-set detection, and OWLv2 scales detection with language-defined categories \citep{liu2024groundingdino,minderer2023owlv2}. Because this reduction already provides vector outputs, UniCounting focuses on proposal grouping and category admission under vector-level counting errors, with a replayable group for every accepted count.

\paragraph{Multi-category counting.}
Reference-less counters typically return one anonymous count \citep{ranjan2022reprpn,hobley2022rcc,xu2023zsc}. ABC123 and OCCAM discover multiple image-specific types without exemplars \citep{hobley2024abc123,spanakis2026occam}, whereas OmniCount names user-specified categories \citep{mondal2025omnicount}. PrACo tests whether text-guided counters reject absent labels, while PrACo++ adds negative-label and distractor protocols and MUCCA provides real multi-category scenes \citep{ciampi2025praco,pacini2026does}. UniCounting complements these directions by using one global vocabulary fixed for the entire run before its first image is processed, emitting every coordinate, and reporting both absent-class false mass and presence F1.

\paragraph{Proposal-based recognition.}
Foundation models support region discovery and semantic matching \citep{kirillov2023sam,ravi2024sam2,radford2021clip,cherti2023openclip,oquab2024dinov2}, but their proposal sets contain duplicates and incomplete views. We isolate the conversion of these redundant proposals into a complete named count vector with group-level evidence.

\section{Method}
\label{sec:method}

\begin{figure*}[t]
\centering
\includegraphics[width=\textwidth]{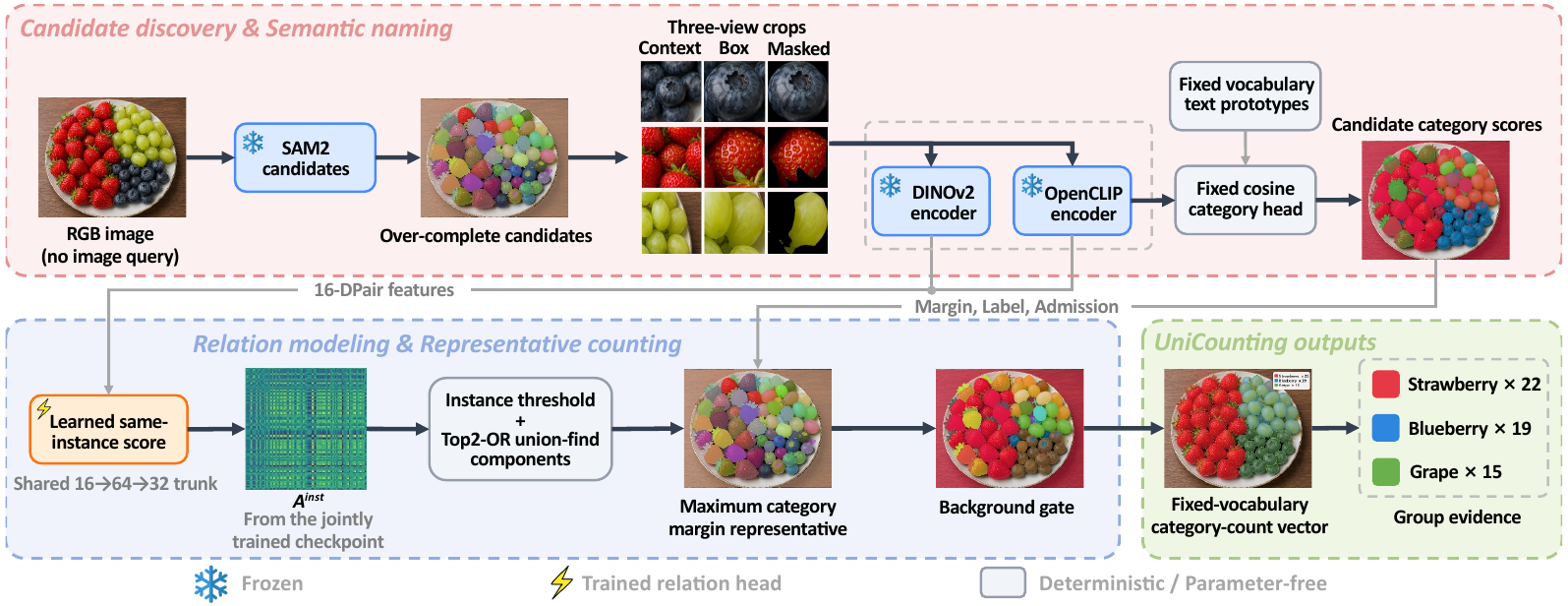}
\caption{UniCounting modules and data flow.
Frozen SAM 2.1 generates over-complete candidates, while frozen DINOv2 and OpenCLIP encode the three proposal views for relation features. The fixed category head uses only the masked- and box-view OpenCLIP embeddings to compute fixed-vocabulary category scores.
The trained relation head supplies the same-instance affinity used by thresholding, top2-OR, and union-find.
Deterministic maximum-category-margin representative selection, raw-cosine labeling, and background-margin admission then produce the count vector and its group evidence. Snowflake, lightning, and gray modules denote frozen, trained, and parameter-free computation, respectively.}
\label{fig:pipeline}
\end{figure*}

\subsection{Task Definition}
\label{sec:task}

For each deployment or evaluation run, a global vocabulary is fixed before any image in that run is processed; no image-dependent category selection is supplied. Let $\mathcal{V}=\{c_1,\ldots,c_K\}$ denote this vocabulary and $I$ an RGB image. We call the task image-query-free because inference receives $I$ but no image-specific text, point, exemplar, target label, or restricted class list. Category names, templates, and text prototypes are fixed together with $\mathcal V$ for the full run. For each category $c$, the model returns a collection $\widehat{\mathcal R}_c$ of admitted proposal components, together with the count vector
\begin{equation}
\widehat{\mathbf n}(I;\mathcal V)
=(\hat n_{c_1},\ldots,\hat n_{c_K}),\qquad
\hat n_c=|\widehat{\mathcal R}_c|,
\label{eq:task}
\end{equation}
where every $G\in\widehat{\mathcal R}_c$ is a connected component of proposals assigned to $c$ and retains a designated representative $k_G^\star$. Thus, $\widehat{\mathcal R}_c$ is a collection of components, not a set of representative proposals. A category is present when $\hat n_c>0$. The predictor cannot read a target class, a per-image class list, or any evaluation annotation. The global vocabulary defines the ontology for the run; it is neither a per-image prompt nor an image-dependent class list.

\subsection{Fixed Category Head}
\label{sec:candidates}

A frozen SAM 2.1 automatic mask generator \citep{ravi2024sam2} produces candidates $\mathcal M=\{m_i\}_{i=1}^{N}$. Each candidate has masked, box, and context crops. Frozen DINOv2 \citep{oquab2024dinov2} and OpenCLIP \citep{radford2021clip,cherti2023openclip} encode all three views for relation prediction. For view $v$, let $\mathbf e_i^v$ denote the L2-normalized OpenCLIP embedding of candidate $i$, and let $\operatorname{norm}(\cdot)$ denote L2 normalization. The category head uses only the masked and box OpenCLIP embeddings. Fixed templates are averaged into a normalized prototype $\mathbf t_c$ for each $c\in\mathcal V$. The proposal feature and its raw category score are computed together as
\begin{align}
\mathbf v_i
&=\operatorname{norm}(0.05\mathbf e_i^{\mathrm{mask}}+0.95\mathbf e_i^{\mathrm{box}}),
\label{eq:proposal_feature}\\
s_{ic}&=\mathbf v_i^\mathsf{T}\mathbf t_c, \qquad c\in\mathcal V.
\label{eq:category_score}
\end{align}
Template lists and taxonomy-specific name aggregation are given in the supplement.
The classifier over $\mathcal V$ has no trainable parameters, no trainable background or unknown class, and no learned rejection gate. A separate fixed bank of seven background prototypes is used only by the admission rule in Section~\ref{sec:admission}. The selected decoder uses the category cosines for representative selection and labeling.

\subsection{Relation Head and Same-Instance Score}
\label{sec:relations}

The relation head is the only trainable module. For each directed pair $i\rightarrow j$, a 16-dimensional feature $\boldsymbol\phi_{i\rightarrow j}$ contains same-view DINO and CLIP cosine similarities, cross-view CLIP similarities, directed log-area ratio, minimum-to-maximum area ratio, two image-area fractions, two SAM 2.1 scores, their absolute difference, and a constant. Each dimension is centered by its relation-training mean and divided by its relation-training standard deviation after clamping that denominator below at $10^{-4}$. Feature 16 is an inactive compatibility placeholder: after standardization it is identically zero and, because the MLP already has affine biases, it does not affect any branch logit. Let $W_1\in\mathbb R^{64\times16}$ and $W_2\in\mathbb R^{32\times64}$ have biases $\mathbf b_1\in\mathbb R^{64}$ and $\mathbf b_2\in\mathbb R^{32}$, and let $D_{0.1}$ denote dropout with rate 0.1. For branch $r\in\mathcal H=\{\mathrm{sem},\mathrm{inst},\mathrm{comp}\}$, let $\mathbf w_r\in\mathbb R^{32}$ and $\beta_r\in\mathbb R$ parameterize its output layer. Let $\sigma(x)=(1+e^{-x})^{-1}$ denote the logistic sigmoid. The directed representation, branch logit, and symmetric same-instance score are
\begin{align}
\mathbf h_{i\rightarrow j}
&=\operatorname{GELU}\!\left(W_2 D_{0.1}
\!\left(\operatorname{GELU}(W_1\boldsymbol\phi_{i\rightarrow j}+\mathbf b_1)\right)+\mathbf b_2\right),
\label{eq:pair_feature}\\
z^r_{i\rightarrow j}
&=\mathbf w_r^\mathsf{T}\mathbf h_{i\rightarrow j}+\beta_r,
\qquad r\in\mathcal H,
\label{eq:relation_logit}\\
a^{\mathrm{inst}}_{ij}
&=\sigma\!\left(\frac{z^{\mathrm{inst}}_{i\rightarrow j}
+z^{\mathrm{inst}}_{j\rightarrow i}}{2}\right).
\label{eq:instance_score}
\end{align}
The two affine trunk layers contain 1,088 and 2,080 parameters, and the three output layers contain 99, giving 3,267 trainable parameters. Primary inference uses only the symmetric same-instance score; the semantic and completeness branches are used only in diagnostic decoder variants.
\subsection{Instance Grouping, Representative Selection, and Labeling}
\label{sec:decoding}

For relation-head run $s$, an unordered edge is eligible when $a^{\mathrm{inst}}_{ij}\geq\tau^{(s)}_{\mathrm{inst}}$, with $\tau^{(s)}_{\mathrm{inst}}$ frozen on inner validation. For each candidate $i$, let $T_i$ contain the neighbor indices $j$ of up to two eligible incident edges with the largest instance scores. The top2-OR edge set is
\begin{equation}
E=\bigl\{\{i,j\}:j\in T_i\ \lor\ i\in T_j\bigr\}.
\label{eq:top2_or}
\end{equation}
An OR rather than mutual selection preserves an edge chosen by either endpoint. Union-find on $E$ produces connected components, including singleton proposals. Each $T_i$ limits direct incident selections but does not bound component size after transitive closure.

Within each component $G$, let $s_{i,(1)}$ and $s_{i,(2)}$ be the largest and second-largest raw category cosines for candidate $i$. UniCounting selects the maximum-category-margin representative and assigns the component the representative's highest-scoring category:
\begin{align}
m_i&=s_{i,(1)}-s_{i,(2)},\qquad
k_G^\star=\arg\max_{i\in G}m_i,
\label{eq:representative}\\
\hat c_G&=\arg\max_{c\in\mathcal V}s_{k_G^\star c}.
\label{eq:component_label}
\end{align}
No member averaging, semantic relation, or learned completeness score enters Eqs.~\ref{eq:top2_or}--\ref{eq:component_label}.

\subsection{Background-Margin Admission}
\label{sec:admission}

Let $\mathcal B_{\mathrm{bg}}=\{b_1,\ldots,b_7\}$ denote the fixed background concepts \emph{background}, \emph{an unknown object}, \emph{other object}, \emph{texture}, \emph{shadow}, \emph{printed text}, and \emph{empty scene}. For $b\in\mathcal B_{\mathrm{bg}}$, the normalized prototype $\mathbf t_b^{\mathrm{bg}}$ is the mean of its normalized fixed-template embeddings; the exact templates are listed in the supplement. For each component $G$, UniCounting computes the admission margin $g_G$, applies its binary admission indicator $A_G$, and accumulates admitted labels:
\begin{align}
g_G&=\max_{c\in\mathcal V}s_{k_G^\star c}
-\max_{b\in\mathcal B_{\mathrm{bg}}}
(\mathbf v_{k_G^\star}^{\mathsf T}\mathbf t_b^{\mathrm{bg}}),
\notag\\[-2pt]
&\hspace{2.5em}A_G=\mathbf{1}[g_G\geq\tau_{\mathrm{bg}}^{(s)}],
\label{eq:background_admission}\\
\hat n_c&=\sum_G A_G\,\mathbf{1}[\hat c_G=c].
\label{eq:count_update}
\end{align}
The run-specific threshold $\tau_{\mathrm{bg}}^{(s)}$ is frozen on inner validation and is a raw-cosine margin rather than a probability. We omit the superscript $(s)$ when unambiguous. A rejected component contributes zero to every category. Its label is not changed, and its count is not redistributed.

\begin{algorithm}[t]
\caption{UniCounting inference with explicit state and outputs.}
\label{alg:unicounting}
\begin{algorithmic}[1]
\REQUIRE Pair scores $a^{\mathrm{inst}}_{ij}$, category scores $s_{ic}$, proposal features $\mathbf v_i$
\REQUIRE Background prototypes $\{\mathbf t_b^{\mathrm{bg}}\}_{b\in\mathcal B_{\mathrm{bg}}}$, thresholds $\tau^{(s)}_{\mathrm{inst}}$ and $\tau_{\mathrm{bg}}^{(s)}$
\ENSURE Count vector $\widehat{\mathbf n}$ and admitted components $\{\widehat{\mathcal R}_c\}_{c\in\mathcal V}$
\FOR{each candidate $i$}
    \STATE $T_i\gets$ the top two $j$ satisfying $a^{\mathrm{inst}}_{ij}\geq\tau^{(s)}_{\mathrm{inst}}$
\ENDFOR
\STATE $E\gets\{\{i,j\}:j\in T_i\ \lor\ i\in T_j\}$ and $U\gets\operatorname{UnionFind}(N)$
\FOR{each $\{i,j\}\in E$}
    \STATE $U.\operatorname{union}(i,j)$
\ENDFOR
\STATE $\widehat{\mathbf n}\gets\mathbf 0_K$ and $\widehat{\mathcal R}_c\gets\varnothing$ for every $c\in\mathcal V$
\FOR{each $G\in U.\operatorname{components}()$}
    \STATE Compute $k_G^\star$, $\hat c_G$, and $g_G$ using Eqs.~\ref{eq:representative}--\ref{eq:background_admission}
    \IF{$g_G\geq\tau_{\mathrm{bg}}^{(s)}$}
        \STATE $\hat n_{\hat c_G}\gets\hat n_{\hat c_G}+1$ and $\widehat{\mathcal R}_{\hat c_G}\gets\widehat{\mathcal R}_{\hat c_G}\cup\{G\}$
    \ENDIF
\ENDFOR
\STATE \textbf{return} $\widehat{\mathbf n},\{\widehat{\mathcal R}_c\}_{c\in\mathcal V}$
\end{algorithmic}
\end{algorithm}

Pair scoring costs $O(N^2)$ before sparsification. The graph stores at most $2N$ endpoint selections, and union-find component construction is near-linear in $N+|E|$. Exact thresholds, index orders, and deterministic tie-breaking rules are specified in the supplement.

\subsection{Supervision and Optimization}
\label{sec:training}

The category head remains fixed, and the relation head is fitted on 210 COCO train2017 images \citep{lin2014coco}, with a disjoint 45-image inner-validation split, denoted inner45, reserved for decoder and threshold selection. A proposal is supervised only when one non-crowd instance accounts for at least 0.8 of its mask, the second-best instance accounts for at most 0.1, and ignore overlap is at most 0.1. Reliable proposal pairs receive same-category and same-instance targets; the completeness target is directional and identifies the proposal with greater ground-truth coverage when the containment and coverage-gap conditions hold. Both directions of each pair are used for training, while Algorithm~\ref{alg:unicounting} reads only the same-instance branch at inference. Pair supports, the weighted training objective, and optimization settings are specified in the supplement.

Selection uses only inner45. It minimizes complete-vector $\ell_1$ for graph construction, representative selection, and labeling, then selects admission by micro presence F1, with vector $\ell_1$ breaking ties. No target-test set enters training or selection.
\begin{table*}[t]
\centering
\setlength{\tabcolsep}{2.0pt}
\begin{tabular*}{\textwidth}{@{\extracolsep{\fill}}l cccccc@{}}
\toprule
Method & Vector $\ell_1\downarrow$ & Present MAE$\downarrow$ & False mass$\downarrow$ & Total MAE$\downarrow$ & Macro-F1$\uparrow$ & Micro-F1$\uparrow$ \\
\midrule
Zero vector & 7.018 & 2.425 & {0.000} & 7.018 & 0.000 & 0.000 \\
Fixed-list detector (OWLv2-All80) & 6.860 & 2.205 & 0.474 & \textbf{5.800} & \textbf{0.472} & 0.415 \\
UniCounting & \textbf{6.485$\pm$0.058} & \textbf{2.129$\pm$0.005} & \textbf{0.325$\pm$0.062} & 5.868$\pm$0.050 & 0.407$\pm$0.007 & \textbf{0.416$\pm$0.005} \\
\midrule
Faster R-CNN (fully supervised ref.) & 3.098 & 0.916 & 0.446 & 2.078 & 0.809 & 0.832 \\
\bottomrule
\end{tabular*}
\caption{Primary complete-vector results on COCO clean500. Every row predicts the same COCO80 vector. OWLv2 receives the full raw COCO80 list identically for every image, while UniCounting uses a fixed taxonomy and no image-specific cue. Its row reports mean$\pm$sample standard deviation over three relation-head seeds; seed-specific thresholds are given in the supplement. The point estimate for UniCounting minus OWLv2 vector $\ell_1$ is $-0.375$ (6.485 versus 6.860). The separated Faster R-CNN row uses full COCO box supervision and is reported only as a supervised reference, not as a like-for-like training-regime comparison.}
\label{tab:coco_decoder}
\end{table*}

\section{Experiments}
\label{sec:experiments}

The experiments ask how UniCounting performs under complete-vector counting, whether the relation-head score improves consolidation under a matched decoder, and which partition changes explain its gains and costs. We then analyze the remaining decoder components and their selection trace. All systems are evaluated under the protocol defined in Section~\ref{sec:task}. OWLv2-All$K$ satisfies the same complete-vector output requirement, although it receives the fixed $K$-class list as an inference input. The Faster R-CNN reference also emits a complete COCO80 vector but uses full COCO box supervision. ABC123 and OCCAM discover anonymous image-specific types, while OmniCount predicts categories requested by a user cue. FSC-147 and CARPK provide only single-coordinate transfer diagnostics under heterogeneous cue protocols, so their results and the full protocol comparison are reported in Supplementary Table~S7 rather than used as primary evidence.

\subsection{Protocols and Metrics}
\label{sec:protocol}

\begin{table*}[t]
\centering
\setlength{\tabcolsep}{2pt}
\begin{tabular*}{\textwidth}{@{\extracolsep{\fill}}l c cccc@{}}
\toprule
Edge/grouping rule & $\tau$ & Vector $\ell_1\downarrow$ & False mass$\downarrow$ & Total MAE$\downarrow$ & Micro-F1$\uparrow$ \\
\midrule
No grouping / singletons & -- & 29.187 & 20.347 & 24.440 & 0.319 \\
Mask containment & 0.005 & 9.935 & 4.697 & 5.020 & 0.427 \\
Mask IoU & 0.005 & 13.649 & 8.265 & 7.965 & 0.374 \\
CLIP cosine & 0.500 & 7.335 & 1.452 & 4.987 & 0.431 \\
DINO cosine & 0.310 & 6.814 & 0.809 & 5.463 & \textbf{0.434} \\
Relation-head score $+$ overlap guard & 0.465 & 9.597 & 4.361 & \textbf{4.635} & 0.414 \\
Relation-head score (shared \(\tau\), controlled) & 0.465 & \textbf{6.483} & \textbf{0.326} & 5.864 & 0.417 \\
\bottomrule
\end{tabular*}
\caption{Controlled edge and grouping ablation on COCO clean500. The two relation-head rows are controlled variants rather than the primary UniCounting configuration: both use a common edge threshold $\tau$=0.465, whereas primary UniCounting uses seed-specific instance thresholds selected on inner45. The table reports end-task metrics after a shared downstream decoder; full pre-admission partition diagnostics are provided in the supplement. Mask IoU is the binary-mask intersection over union; the overlap guard instead adds a positive-intersection constraint to the relation-head score. Every row is the arithmetic mean of seed-specific point metrics for seeds 17, 42, and 73. For containment, mask IoU, CLIP, DINO, relation-head, singleton, and overlap-guard rows, the candidate pool, pair universe, graph construction, representative and labeling rules, and run-matched background-margin admission are fixed; only the edge score, common edge threshold, or stated grouping control changes.}
\label{tab:edge_ablation}
\end{table*}

\paragraph{Model selection and system variants.}
All decoder choices and thresholds are fixed on a COCO inner-validation split before target-test evaluation. The unqualified name \emph{UniCounting} refers only to Algorithm~\ref{alg:unicounting}.

\paragraph{Controlled edge and grouping protocol.}
We replace only the symmetric edge score with mask containment, mask IoU, CLIP cosine, DINO cosine, or the relation-head score, while fixing the candidate pool, pair universe, top2-OR graph construction, representative rule, labeling rule, and run-matched background-margin admission. Singleton and overlap-guard controls retain the same downstream decoder. Score-specific thresholds are selected only on inner45; exact values and tie-breaking rules are given in the supplement.

\paragraph{Fixed-list detector baseline.}
OWLv2-All80 receives the same list of 80 raw COCO category names for every image, applies classwise NMS, and counts retained detections. Its score threshold is calibrated only on inner45; checkpoint, preprocessing, search-grid, and tie-breaking details are given in the supplement. Clean500 remains untouched.

\paragraph{Fully supervised detector reference.}
We report torchvision Faster R-CNN with a ResNet-50-FPN v2 backbone and official COCO-trained weights \citep{ren2015fasterrcnn}. Classwise NMS and an inner45-selected score threshold produce a complete COCO80 count vector. Since the model uses COCO box supervision, it provides supervised context rather than a like-for-like comparison to UniCounting.

\paragraph{Primary complete-vector evaluation.}
COCO clean500 is a deterministic \COCOImages{}-image subset of val2017 \citep{lin2014coco}. We rank eligible images by file name, exclude identities from the separate 60-image diagnostic split clean60 defined in the supplement, and take the first 500 without reading annotations or model outputs. Its image hashes are disjoint from relation-head training and inner45 validation, and all thresholds are frozen before preprocessing. Let $\{(I_i,\mathbf y_i)\}_{i=1}^{M}$ denote the evaluation set, where $\mathbf y_i=(y_{i1},\ldots,y_{iK})\in\mathbb N_0^K$ is the ground-truth count vector and $\hat{\mathbf y}_i=\widehat{\mathbf n}(I_i;\mathcal V)$ its prediction; here $\mathbb N_0$ denotes the nonnegative integers. For clean500, $M=\COCOImages{}$ and $K=\COCOClasses{}$. Each image contains a mean/median of \COCOPresentMean{}/\COCOPresentMedian{} present categories and \COCOObjectMean{}/\COCOObjectMedian{} annotated objects; \COCOZeroRowRate{}\% of image--class cells are zero, and the largest per-class count is \COCOMaxClassCount{}.

We report mean vector $\ell_1$ error $M^{-1}\sum_i\lVert\hat{\mathbf y}_i-\mathbf y_i\rVert_1$ and present-cell MAE over cells with $y_{ic}>0$. Absent-class false mass is $M^{-1}\sum_i\sum_{c:y_{ic}=0}\hat y_{ic}$, and total-count MAE is $M^{-1}\sum_i|\sum_c\hat y_{ic}-\sum_c y_{ic}|$. We also report macro and micro presence F1, where presence means a positive count; macro-F1 averages represented classes, whereas micro-F1 pools all image--category decisions. Full support and bootstrap rules are given in the supplement.

\paragraph{Transfer evaluations.}
OmniCount-sub is an internally constructed, exact-duplicate-screened 2,000-image holdout sampled from the OmniCount training/validation resources rather than the official test split \citep{mondal2025omnicount}. It uses the 191-category taxonomy released publicly with OmniCount; we use PF191 as shorthand for this public, fixed output space. We therefore treat it as a project holdout, not an official benchmark. Its labels are not used to train the relation head, select the decoder, or tune admission, and we report vector $\ell_1$ error and pooled micro presence F1.

FSC-147 provides all 1,190 test images but only one annotated target category per image \citep{ranjan2021famnet}. UniCounting predicts the complete 147-dimensional vector without target metadata, after which the evaluator extracts the target coordinate for MAE and RMSE. CARPK is evaluated analogously on its 459 test images as zero-shot transfer to the fixed \emph{car} coordinate, without fine-tuning \citep{hsieh2017carpk}. Because FSC-147 and OmniCount-sub share 26 source-image hashes, we report them separately and do not treat them as independent transfer confirmations.

\paragraph{Statistical reporting and reproducibility.}
Unless stated otherwise, relation-head rows report mean and sample standard deviation over three relation seeds. Paired 95\% bootstrap intervals are computed from saved prediction vectors; resampling, seed aggregation, macro-support, and random-seed details are provided in the supplement.

\begin{figure}[t]
\centering
\includegraphics[width=0.47\textwidth]{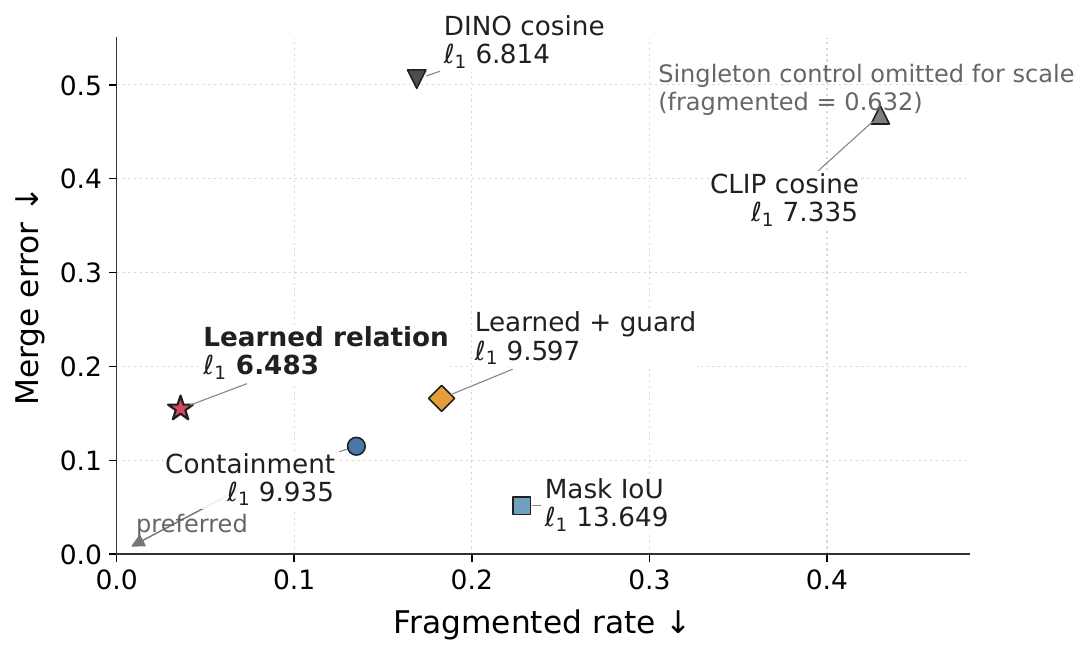}
\caption{Fragmentation--merge trade-off under the matched decoder on COCO clean500. Each point represents an edge or grouping rule; the horizontal and vertical axes report the fragmented-instance rate and component merge error, respectively, so lower-left is preferred. Point annotations report the resulting complete-vector $\ell_1$ error after the shared downstream decoder. The relation-head score produces the lowest fragmentation and vector error among the tested edge scores, while incurring more merges than the conservative mask-containment and mask-IoU rules.}
\label{fig:tradeoff}
\end{figure}

\subsection{Implementation Details}
\label{sec:implementation}

We use frozen SAM 2.1 for proposals, DINOv2 for relation features, and OpenCLIP for category features. The relation head scores all $N(N-1)/2$ proposal pairs before graph sparsification, giving $O(N^2)$ relation-scoring cost. Exact checkpoints, candidate filtering thresholds, crop construction, preprocessing, and deterministic inference settings are specified in the supplement.

\subsection{How does UniCounting perform under complete-vector counting?}
\label{sec:main_results}

Table~\ref{tab:coco_decoder} reports complete-vector performance on COCO clean500. UniCounting is numerically lower than OWLv2-All80 in vector $\ell_1$ (6.485 versus 6.860), present-cell MAE (2.129 versus 2.205), and absent-class false mass (0.325 versus 0.474), while their micro-F1 values are nearly identical (0.416 versus 0.415). OWLv2 has lower total-count MAE (5.800 versus 5.868) and higher macro-F1 (0.472 versus 0.407). We therefore describe the 0.375 vector-$\ell_1$ gap as a point-estimate difference and make no significance claim. The matched-decoder study below separates the contribution of proposal consolidation from the remaining system components.

The fully supervised Faster R-CNN reference reaches 3.098 vector $\ell_1$ and 0.832 micro-F1. This gap indicates substantial headroom when COCO box supervision is available, but it does not isolate the contribution of proposal consolidation because the detector and UniCounting use different training supervision.

The standalone OmniCount-sub complete-vector comparison is reported in Supplementary Table~S5, while FSC-147 and CARPK single-coordinate diagnostics are reported in Supplementary Table~S7. These results provide transfer context rather than evidence for the grouping mechanism.

\subsection{Does the relation-head score improve proposal consolidation under a matched decoder?}
\label{sec:edge_ablation}

Table~\ref{tab:edge_ablation} isolates the edge score under a shared downstream decoder. The relation-head score obtains the lowest vector $\ell_1$ and false mass among the tested scores. Its paired vector-$\ell_1$ and false-mass intervals against containment, mask IoU, CLIP, and DINO exclude zero. DINO has higher micro-F1, and the overlap guard has lower total-count MAE. Thus, the relation-head score improves the two complete-vector error measures targeted by this comparison, but not every end-task metric.

\subsection{Why does it help, and what does it sacrifice?}
\label{sec:grouping_diagnostics}

Figure~\ref{fig:tradeoff} summarizes the mechanism underlying Table~\ref{tab:edge_ablation}. Mask IoU and containment suppress merges more conservatively but leave substantially more instances fragmented. In contrast, the relation-head score shifts the partition toward aggressive consolidation, obtaining the lowest fragmented rate and complete-vector $\ell_1$ error among the tested edge scores, while accepting a moderate merge-error increase. It also attains a Pair F1 of 0.694, purity of 0.925, and representative IoU of 0.215; formal definitions and the complete partition diagnostics are provided in Supplementary Table~S8.

\begin{table}[t]
\centering
\setlength{\tabcolsep}{1pt}
\begin{tabular}{llcc}
\toprule
Stage & Selected variant & Vector $\ell_1\downarrow$ & Micro-F1$\uparrow$ \\
\midrule
Graph & same-instance edges$\star$& $6.867{\pm}0.06$ & $0.417{\pm}0.005$ \\
Represent. & max category margin & $6.837{\pm}0.09$ & $0.423{\pm}0.004$ \\
Labeling & raw cosines & $6.837{\pm}0.09$ & $0.423{\pm}0.004$ \\
Admission & background-margin& $6.793{\pm}0.05$ & $0.426{\pm}0.007$ \\
\bottomrule
\end{tabular}

\caption{Decoder choices selected on inner45. $\star$ refers to all eligible/top2-OR. Each stage carries forward the choices selected at preceding stages. The table is a staged selection trace rather than a cumulative-improvement analysis. The complete candidate matrix and frozen target-set results are reported in the supplement.}
\label{tab:decoder_selection}
\end{table}

\subsection{Decoder Component Ablation and Selection Analysis}
\label{sec:decoder_ablation}

\begin{figure}[t]
\centering
\includegraphics[width=0.47\textwidth]{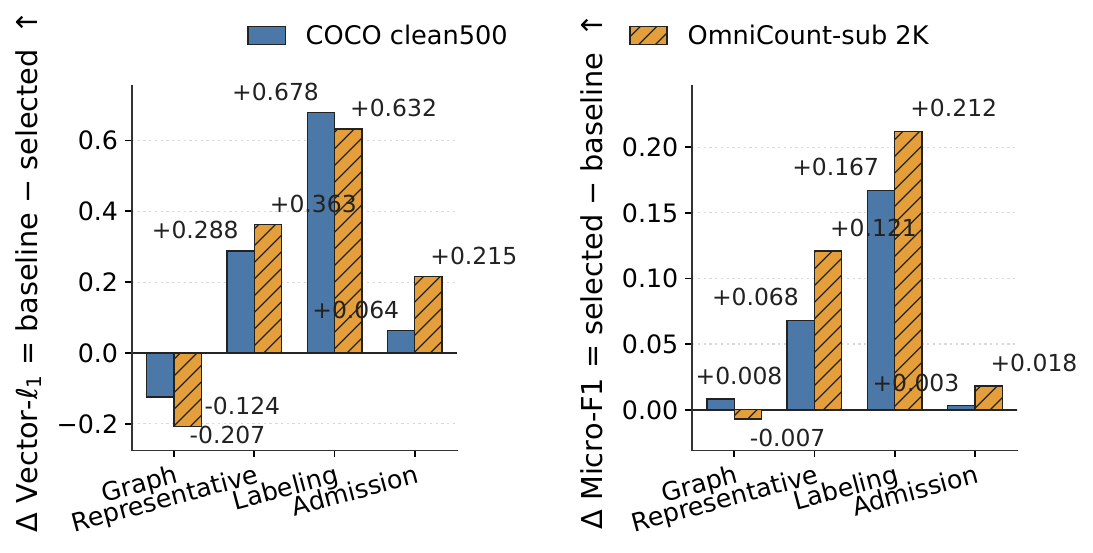}
\caption{Stage-wise effects of the decoder choices selected on inner45 when audited on the frozen clean500 and OmniCount-sub target sets. Panel (a) reports vector-$\ell_1$ improvement, defined as the baseline error minus the selected-variant error; panel (b) reports micro-F1 improvement, defined as the selected-variant score minus the baseline score. Positive values therefore denote improvement in both panels. Baselines are all-edges grouping, maximum-area representation, component-mean labeling, and no admission, respectively. Representative selection and labeling provide the largest consistent gains, whereas the effect of top2-OR sparsification is metric- and dataset-dependent.}
\label{fig:decoder_effects}
\end{figure}

Table~\ref{tab:decoder_selection} records the staged inner45 selection trace, with each row carrying forward the choices selected at preceding stages. Figure~\ref{fig:decoder_effects} separates the contribution of each selected decoder choice on the two frozen target sets. Maximum-category-margin representation and representative raw-cosine labeling consistently improve both vector $\ell_1$ and micro-F1, with labeling producing the largest effect. Background-margin admission provides a smaller additional gain. In contrast, top2-OR sparsification exhibits metric- and dataset-dependent behavior, supporting selection of the complete pipeline on inner45 rather than retrospective target-set tuning. The complete audit in Supplementary Table~S9 further shows that the semantic constraint consistently degrades performance.

\section{Discussion and Limitations}
\label{sec:limitations}

The complete-vector evidence is limited to the 500-image COCO evaluation and the project holdout OmniCount-sub, and we do not evaluate MUCCA or PrACo++. clean500 is disjoint from development data and selected without annotations, but remains a subset of COCO val2017, with unknown foundation-model pretraining overlap. The controlled ablation isolates five edge scores and two grouping controls under one decoder. The relation-head score reduces vector error and false mass relative to tested alternatives, but does not minimize total-count MAE; its merge error is higher than mask IoU and containment, while lower than CLIP and DINO. The candidate pool recalls only 62.4\% of non-ignore instances at mask IoU 0.50. Since representative selection and admission maximize over $\mathcal V$, changing vocabularies shifts score distributions, and reusing COCO-calibrated thresholds may impair transfer calibration.

Trainable capacity is limited to the 3,267-parameter relation head, but inference still runs frozen SAM 2.1, DINOv2, and OpenCLIP components with quadratic pair scoring. Deployment should further address privacy, ontology omissions, and systematic category errors.

\section{Conclusion}
\label{sec:conclusion}

UniCounting studies image-query-free multi-category counting from the perspective of proposal consolidation. Instead of directly mapping redundant segmentation proposals to category counts, it explicitly models proposal identity, forms instance-level components, and produces category-count outputs with traceable group evidence. A 3,267-parameter category-shared relation head improves proposal partition quality under a controlled decoder comparison, reducing vector-level counting error relative to several similarity-based alternatives.
Our analysis also reveals an inherent trade-off in proposal consolidation: more aggressive grouping reduces fragmentation but may increase instance merging errors. These findings suggest that reliable multi-category counting requires not only semantic recognition but also explicit control over instance identity. Future work should improve proposal recall and develop more adaptive mechanisms for balancing merge and fragmentation errors.

\section*{Acknowledgments}

This work was supported by the National Natural Science Foundation (No.~62503505) and the Hunan Natural Science Foundation (No.~2025JJ60424).

\bibliography{aaai2027}

\end{document}


\maketitle

This supplement specifies the implementation choices, evaluator definitions, and supporting experiments for the method in the main paper.

\section{Reproducibility Protocol}
\label{supp:manifest}

\begin{center}
\setlength{\tabcolsep}{3pt}
\begin{tabular}{@{}p{0.43\columnwidth}p{0.52\columnwidth}@{}}
\toprule
Field & Configuration \\
\midrule
Training seeds & 17, 42, 73 \\
COCO \citep{lin2014coco} release and split & val2017 clean500; fixed SHA256-ranked identities \\
Threshold-selection split & disjoint 45-image COCO inner-validation split (inner45) \\
Bootstrap repetitions / seed & 10,000 / 20260725 \\
\bottomrule
\end{tabular}
\captionof{table}{Reproducibility settings for training, selection, and evaluation.}
\label{tab:manifest}
\end{center}

\paragraph{Development-split provenance.}
The training code consumes two fixed COCO train2017 role assignments stored in the source registry: 210 images marked \texttt{relation\_train} and 45 images marked \texttt{inner\_validation}. It does not generate either role by a random draw, a hash rule, or metric-based filtering. The sets contain 210 and 45 unique identities, respectively, with no overlap. SHA-256 of the numerically sorted, newline-delimited, 12-digit image IDs is \texttt{8f5728c98bf1} for train210 and \texttt{06744dca51b6} for inner45. Train210 contains 11,341 candidates, 3,262 reliable proposals, and 45,508 reliable unordered pairs; inner45 covers 60 of the 80 COCO categories.

\paragraph{COCO clean60 diagnostic set.}
Clean60 is the fixed 60-image COCO val2017 identity list in the released split file (identity SHA-256 \texttt{f1fcbdcf}). Because its construction does not specify a random-sampling formula, we treat clean60 as a diagnostic set and make no population-level representativeness claim. It is not used for relation-head training, decoder selection, or threshold selection. Its image hashes are disjoint from the 210-image relation-training set and the 45-image inner-validation set, both drawn from train2017, and clean500 is constructed after excluding all clean60 identities. The COCO80 targets contain 340 non-ignore instances from 54 categories and 155 positive image--category cells. Per image, the mean/median non-ignore instance count is 5.667/4 and the mean/median number of present categories is 2.583/2; the largest single-category count is 13, and 96.771\% of the $60\times80$ cells are zero.

\paragraph{Image-query-free protocol.}
UniCounting follows the run-level vocabulary protocol of Section~3.1. No per-image category cue is supplied, and FSC-147 target coordinates are extracted only after the full fixed-vocabulary prediction is produced.

\paragraph{FSC-147 annotation boundary.}
FSC-147 specifies one benchmark target but not every countable category in a scene.

\begin{figure*}[t]
\centering
\includegraphics[width=0.78\textwidth]{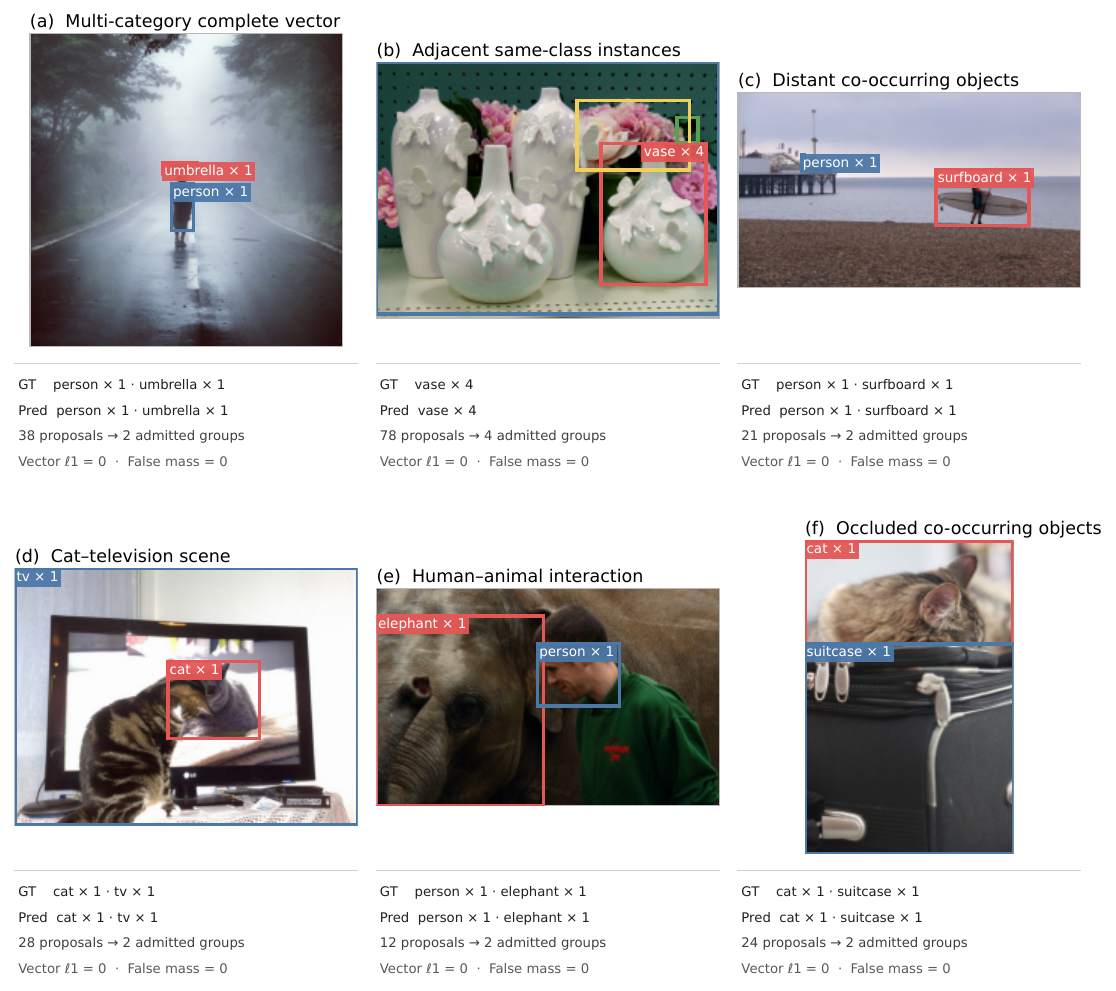}
\caption{Selected successful complete-vector predictions on COCO clean500. All examples use the primary UniCounting decoder with the fixed seed-17 checkpoint and thresholds selected on inner45. Colored contours indicate the representatives of admitted proposal components, and the accompanying summaries compare all nonzero ground-truth and predicted coordinates. The examples were filtered using fixed correctness and consolidation-complexity criteria and then selected for visual legibility and diversity; they are intended to illustrate successful behaviors rather than estimate average performance. Ground-truth annotations are used only for visualization and evaluation.}
\label{fig:successful_predictions_supp}
\end{figure*}

\section{Implementation Details}
\label{supp:implementation}

\subsection{Computing Environment}

All experiments were executed in the computing environment summarized below. The CUDA version supported by the NVIDIA driver is distinguished from the CUDA version against which PyTorch was compiled.

\begin{table}[t]
\centering
\setlength{\tabcolsep}{4pt}
\begin{tabular}{p{0.36\columnwidth}p{0.55\columnwidth}}
\toprule
Setting & Value \\
\midrule
SAM 2.1 \citep{ravi2024sam2} checkpoint & \texttt{sam2.1-hiera-small}, SHA-256 prefix \texttt{0a4067b1} \\
Input resolution & checkpoint-native preprocessing \\
Points per side / crop layers & 24 / 0 \\
Predicted-IoU threshold & 0.7 \\
Stability threshold & 0.8 \\
Box NMS threshold & 0.7 \\
Minimum / maximum mask area fraction & $10^{-4}$ / 0.95 \\
Near-duplicate rule & descending predicted IoU; discard if box IoU $>0.5$ and mask IoU $>0.9$ against an already-retained mask \\
Minimum box side / tiling & 4 pixels / none \\
\bottomrule
\end{tabular}
\caption{Automatic mask generation and filtering.}
\label{tab:sam_config}
\end{table}

{\setlength{\tabcolsep}{2.5pt}
\begin{center}
\begin{tabular}{@{}clcl@{}}
\toprule
Feature & Definition & Feature & Definition \\
\midrule
1 & $\cos(\mathbf d_i^m,\mathbf d_j^m)$ & 9 & $\operatorname{clip}(\log(A_i/A_j),-5,5)/5$ \\
2 & $\cos(\mathbf d_i^b,\mathbf d_j^b)$ & 10 & $\min(A_i,A_j)/\max(A_i,A_j)$ \\
3 & $\cos(\mathbf d_i^x,\mathbf d_j^x)$ & 11 & $\operatorname{clip}(A_i/A_I,0,1)$ \\
4 & $\cos(\mathbf c_i^m,\mathbf c_j^m)$ & 12 & $\operatorname{clip}(A_j/A_I,0,1)$ \\
5 & $\cos(\mathbf c_i^b,\mathbf c_j^b)$ & 13 & $q_i$ \\
6 & $\cos(\mathbf c_i^x,\mathbf c_j^x)$ & 14 & $q_j$ \\
7 & $\cos(\mathbf c_i^m,\mathbf c_j^b)$ & 15 & $|q_i-q_j|$ \\
8 & $\cos(\mathbf c_i^b,\mathbf c_j^m)$ & 16 & $1$ \\
\bottomrule
\end{tabular}
\end{center}}

\begin{table}[t]
\centering
\setlength{\tabcolsep}{4pt}
\begin{tabular}{p{0.36\columnwidth}p{0.55\columnwidth}}
\toprule
Setting & Value \\
\midrule
GPU / memory & NVIDIA GeForce RTX 4090 / 24,564 MiB (approximately 24 GB) \\
CPU & Intel Core i9-14900KF\\
System memory & 62 GiB available (approximately 64 GB physical memory) \\
Operating system & Ubuntu 24.04.4 LTS under WSL2 \\
Linux kernel & \texttt{6.6.114.1-}\newline\texttt{microsoft-standard-WSL2} \\
Python & 3.12.3 \\
NVIDIA driver / supported CUDA & 591.86 / 13.1 \\
PyTorch / build CUDA & \texttt{2.12.0+cu130} / 13.0 \\
torchvision / cuDNN & \texttt{0.27.0+cu130} / 9.2.0 \\
\bottomrule
\end{tabular}
\caption*{Hardware and software environment used for the experiments.}
\end{table}

\subsection{Candidate Generation}

The same 24-point grid, zero crop layers, and no-tiling path is applied to every image. Candidates are processed greedily in descending predicted IoU for near-duplicate removal; exact predicted-IoU ties preserve the mask generator's output order. The formal decoder contains no density, resolution, count-bin, or ground-truth-dependent router.

\subsection{Category and Relation Heads}

COCO category prototypes average seven templates: \texttt{a photo of a \{\}.}, \texttt{a photo of the \{\}.}, \texttt{a close-up photo of a \{\}.}, \texttt{a cropped photo of a \{\}.}, \texttt{a photo of one \{\}.}, \texttt{an image of a \{\}.}, and \texttt{a \{\} in the scene.}. FSC-147 and OmniCount use \texttt{a photo of \{phrase\}.}, \texttt{a close-up photo of \{phrase\}.}, \texttt{multiple \{phrase\}.}, \texttt{a group of \{phrase\}.}, and \texttt{cropped \{phrase\}.}. FSC-147 combines canonical, alias, and description prototype groups with weights 0.50/0.25/0.25; OmniCount uses canonical taxonomy names only. For each of the seven background concepts in $\mathcal B_{\mathrm{bg}}$, $\mathbf t_b^{\mathrm{bg}}$ averages the normalized embeddings of \texttt{a photo of \{\}.}, \texttt{a cropped photo of \{\}.}, and \texttt{\{\} in the scene.}.

\paragraph{Directed relation features.}
Let $\mathbf d_i^v$ and $\mathbf c_i^v$ be the L2-normalized DINOv2 \citep{oquab2024dinov2} and OpenCLIP \citep{radford2021clip,cherti2023openclip} embeddings for view $v\in\{m,b,x\}$ (masked, box, and context), let $A_i$ be mask area in pixels, and let $A_I$ be image area. We define $q_i$ as the SAM 2.1 predicted-IoU score returned for candidate $i$. The stability score is used only for the 0.8 candidate-filtering threshold in Table~\ref{tab:sam_config} and is not part of $q_i$. With $\operatorname{cos}(\mathbf a,\mathbf b)=\mathbf a^\mathsf{T}\mathbf b$, the feature order for $i\rightarrow j$ is fixed as follows.
The reverse feature $\boldsymbol\phi_{j\rightarrow i}$ is constructed by exchanging $i$ and $j$ in every row; in particular, Features 7--8, 11--12, and 13--14 exchange positions, while Feature 9 changes sign before clipping. Each dimension is centered by its relation-training mean and divided by its relation-training standard deviation after clamping that denominator below at $10^{-4}$. Feature 16 is therefore identically zero after standardization. It is retained only as an inactive compatibility placeholder and does not affect the logits; constant offsets are already supplied by the MLP biases.

\subsection{Pseudo-Label Construction}

For candidate $m_i$ and non-ignore ground-truth instance $G_u$, define intersection-over-union, candidate purity, and ground-truth coverage as
\begin{equation}
\begin{aligned}
\operatorname{IoU}_{iu}&=\frac{|m_i\cap G_u|}{|m_i\cup G_u|},
&\pi_{iu}&=\frac{|m_i\cap G_u|}{|m_i|},\\
\kappa_{iu}&=\frac{|m_i\cap G_u|}{|G_u|}. &&
\end{aligned}
\end{equation}
The matched instance is $u_i=\arg\max_u\pi_{iu}$, with the lower stored GT index breaking exact ties, and $\pi_{i,(2)}$ denotes the second-largest non-ignore purity. Candidate $i$ is reliable exactly when $\pi_{iu_i}\geq0.8$, $\pi_{i,(2)}\leq0.1$, and its total overlap with ignore or crowd regions is at most $0.1|m_i|$. Same-category and same-instance pair targets compare the category and instance identities of reliable candidates. A directed completeness positive additionally requires both reliable proposals to match the same instance, containment $|m_i\cap m_j|/\min(|m_i|,|m_j|)\geq0.9$, and a GT-coverage advantage of at least 0.2 for the larger-coverage proposal. No additional proposal-status classes or heuristic proposal labels are introduced.

\begin{center}
\centering
\small
\setlength{\tabcolsep}{4pt}
\begin{tabular}{p{0.36\columnwidth}p{0.55\columnwidth}}
\toprule
Setting & Value \\
\midrule
DINOv2 checkpoint & \texttt{facebook/dinov2-small}, revision prefix \texttt{ed25f3a} \\
CLIP checkpoint & \texttt{ViT-B-32}\newline\texttt{laion2b\_s34b\_b79k}, snapshot prefix \texttt{1a25a44} \\
Masked background fill & black \\
Context expansion / CLIP resolution & half a box width and height / $224\times224$ \\
Proposal feature weights & 0.05 masked + 0.95 box, then L2 normalization \\
Activation / normalization / dropout & GELU / training-set standardization / 0.1 \\
Relation feature dimension & 16 directed pair features \\
Relation MLP dimensions / parameters & $16\!\rightarrow\!64\!\rightarrow\!32$, three biased $32\!\rightarrow\!1$ outputs / $1{,}088+2{,}080+99=3{,}267$ \\
Pair rows / pruning & both directions of every reliable unordered pair / none before scoring \\
\bottomrule
\end{tabular}
\captionof{table}{Feature extraction and trainable-head architecture.}
\label{tab:head_config}
\end{center}

\begin{table}[t]
\centering
\setlength{\tabcolsep}{3pt}
\begin{tabular}{lccc}
\toprule
Edge score & threshold & vector $\ell_1\downarrow$ & micro-F1$\uparrow$ \\
\midrule
Mask containment & 0.005 & 12.5333 & 0.4289 \\
Mask IoU & 0.005 & 15.8222 & 0.3920 \\
CLIP cosine & 0.500 & 7.2444 & 0.4632 \\
DINO cosine & 0.310 & 7.1185 & 0.4160 \\
Relation-head score & 0.465 & 6.7930 & 0.4260 \\
\bottomrule
\end{tabular}
\caption{Frozen threshold selection for the controlled edge-score study in inner45. Mask containment is $|m_i\cap m_j|/\min(|m_i|,|m_j|)$, whereas mask IoU is $|m_i\cap m_j|/|m_i\cup m_j|$. CLIP and DINO use cosine similarity between normalized $0.05$ masked-view plus $0.95$ box-view features.}
\label{tab:edge_thresholds_supp}
\end{table}

\begin{table}[t]
\centering
\setlength{\tabcolsep}{10pt}
\begin{tabular}{lcc}
\toprule
Method & Vector $\ell_1\downarrow$ & Micro-F1$\uparrow$ \\
\midrule
Zero vector & {6.447} & 0.000 \\
UniCounting & 6.715$\pm$0.123 & 0.283$\pm$0.011 \\
\bottomrule
\end{tabular}
\caption{OmniCount-sub complete-vector diagnostic. Both rows predict the same 191-dimensional vector. OmniCount-sub is a project holdout used only as a transfer diagnostic, not as an official benchmark.}
\label{tab:omni_vector_supp}
\end{table}

\begin{table}[t]
\centering
\setlength{\tabcolsep}{1pt}
\begin{tabular}{lrrrrrr}
\toprule
Method / seed & TP & FP & FN & Micro-P & Micro-R & Micro-F1 \\
\midrule
OWLv2-All80 & 432 & 202 & 1015 & 0.681 & 0.299 & 0.415 \\
UniCounting / 17 & 429 & 187 & 1,018 & 0.696 & 0.296 & 0.416 \\
UniCounting / 42 & 418 & 165 & 1,029 & 0.717 & 0.289 & 0.412 \\
UniCounting / 73 & 420 & 125 & 1,027 & 0.771 & 0.290 & 0.422 \\
\bottomrule
\end{tabular}
\caption{Pooled image--category presence diagnostics on COCO clean500.
TP, FP, and FN are reported for each prediction vector; precision, recall, and F1 are computed from those pooled counts.
Its low micro-F1 is primarily associated with low micro recall rather than excessive false-positive presence.}
\label{tab:presence_diagnostics_supp}
\end{table}

\begin{table*}[!t]
\centering
\setlength{\tabcolsep}{2.0pt}
\begin{tabular*}{\textwidth}{@{\extracolsep{\fill}}lcccccc@{}}
\toprule
Grouping rule & Pair F1$\uparrow$ & Purity$\uparrow$ & Merge error$\downarrow$ & Comp./instance$\downarrow$ & Fragmented$\downarrow$ & Rep. IoU$\uparrow$ \\
\midrule
No grouping / singletons & 0.0000 & 1.0000 & 0.0000 & 4.0770 & 0.6320 & 0.1680 \\
Mask containment & 0.6910 & 0.9090 & 0.1150 & 1.4150 & 0.1350 & 0.2020 \\
Mask IoU & 0.4984 & \textbf{0.9558} & \textbf{0.0515} & 1.8104 & 0.2282 & 0.1758 \\
CLIP cosine & 0.5040 & 0.6740 & 0.4680 & 1.6320 & 0.4300 & 0.1450 \\
DINO cosine & 0.6270 & 0.6510 & 0.5060 & 1.1950 & 0.1690 & 0.1760 \\
Relation-head score $+$ overlap guard & 0.6820 & 0.8500 & 0.1660 & 1.4180 & 0.1830 & 0.2030 \\
Relation-head score & \textbf{0.6940} & 0.9250 & 0.1550 & \textbf{1.0370} & \textbf{0.0360} & \textbf{0.2150} \\
\bottomrule
\end{tabular*}
\caption{Full pre-admission partition diagnostics for the controlled edge and grouping ablation on COCO clean500. Rows follow Table~2 of the main paper. Every value is the arithmetic mean of seed-specific point metrics for seeds 17, 42, and 73. The candidate pool, pair universe, graph construction, representative rule, and labeling rule are fixed; only the edge score, common edge threshold, or stated grouping control changes. Admission is not applied for these partition diagnostics.}
\label{tab:edge_grouping_supp}
\end{table*}

\begin{table*}[!t]
\centering
\setlength{\tabcolsep}{2.0pt}
\textbf{(a) Protocol comparison}\\[2pt]
\begin{tabular*}{\textwidth}{@{\extracolsep{\fill}}>{\raggedright\arraybackslash}p{0.10\textwidth}>{\raggedright\arraybackslash}p{0.19\textwidth}>{\raggedright\arraybackslash}p{0.19\textwidth}>{\raggedright\arraybackslash}p{0.21\textwidth}>{\centering\arraybackslash}p{0.15\textwidth}@{}}
\toprule
Method & Inference-time category input & Category space & Output form & Direct complete-vector comparison \\
\midrule
ABC123 \citep{hobley2024abc123} & None & Image-specific discovered types & Anonymous multi-type counts & No \\
OCCAM \citep{spanakis2026occam} & None & Image-specific discovered types & Anonymous multi-type counts & No \\
OmniCount & Image-dependent category or point cue & Queried semantic categories & Named queried counts & Partial \\
OWLv2 (All$K$) & Fixed $K$-class list & Run-level global named vocabulary & Complete $K$-vector after detection & Yes \\
UniCounting & None & Run-level global named vocabulary & Complete $K$-vector with group evidence & Yes \\
\bottomrule
\end{tabular*}
\textbf{(b) Cross-protocol single-coordinate diagnostics}\\[2pt]
\begin{tabular*}{\textwidth}{@{\extracolsep{\fill}}>{\raggedright\arraybackslash}p{0.26\textwidth}>{\raggedright\arraybackslash}p{0.15\textwidth}cccc@{}}
\toprule
& & \multicolumn{2}{c}{FSC-147} & \multicolumn{2}{c}{CARPK \citep{hsieh2017carpk}} \\
\cmidrule(lr){3-4}\cmidrule(lr){5-6}
Method & Per-image cue & MAE$\downarrow$ & RMSE$\downarrow$ & MAE$\downarrow$ & RMSE$\downarrow$ \\
\midrule
CountGD \citep{amini2024countgd} & Image-dependent text/exemplar & 5.74 & 24.09 & 3.68 & 5.17 \\
OmniCount \citep{mondal2025omnicount} & Image-dependent point prompt & 19.24 & 115.27 & 10.66 & 13.15 \\
OCCAM-S \citep{spanakis2026occam} & none & 16.92 & 110.83 & 10.06 & 13.81 \\
UniCounting & none & 26.409$\pm$0.008 & 120.717$\pm$0.006 & 4.057 & 5.509 \\
\bottomrule
\end{tabular*}
\caption{Protocol comparison and cross-protocol transfer diagnostics. UniCounting follows the run-level vocabulary protocol defined in Section~3.1 of the main paper. Panel (a) distinguishes the inference-time category input and output space of each method; direct comparability means evaluation against a complete vector over a run-level global named vocabulary. ABC123 is omitted from panel (b) because its reported FSC-147 transfer result uses only images containing fewer than 300 objects rather than the full test split. OCCAM-S is the single-class OCCAM configuration reported by \citet{spanakis2026occam}. UniCounting's FSC-147 values are mean$\pm$sample standard deviation over three relation-head seeds. CARPK is reported from a single deterministic run; seed variability was not evaluated. These are single-coordinate transfer diagnostics rather than complete-vector evaluations; methods differ in their inference-time category inputs.}
\label{tab:protocol_transfer_supp}
\end{table*}

\begin{table*}[!t]
\centering
\setlength{\tabcolsep}{1.0pt}
\begin{tabular*}{\textwidth}{@{\extracolsep{\fill}}l l cc cc cc@{}}
\toprule
& & \multicolumn{2}{c}{Inner45} & \multicolumn{2}{c}{COCO clean500} & \multicolumn{2}{c}{OmniCount-sub 2K} \\
\cmidrule(lr){3-4}\cmidrule(lr){5-6}\cmidrule(lr){7-8}
Stage & Variant & $\ell_1\downarrow$ & Micro-F1$\uparrow$ & $\ell_1\downarrow$ & Micro-F1$\uparrow$ & $\ell_1\downarrow$ & Micro-F1$\uparrow$ \\
\midrule
\multirow{3}{*}{Graph} & hard semantic $\land$ instance& 32.978$\pm$0.868 & 0.300$\pm$0.003 & 30.597$\pm$1.040 & 0.291$\pm$0.003 & 26.553$\pm$1.073 & 0.144$\pm$0.003 \\
& same-instance edges, all & 6.889$\pm$0.000 & 0.365$\pm$0.000 & 6.686$\pm$0.000 & 0.342$\pm$0.000 & 6.769$\pm$0.000 & 0.257$\pm$0.000 \\
& same-instance edges, top2$\star$& 6.867$\pm$0.059 & 0.417$\pm$0.005 & 6.810$\pm$0.010 & 0.350$\pm$0.002 & 6.976$\pm$0.009 & 0.250$\pm$0.002 \\
\midrule
\multirow{3}{*}{Represent.} & maximum area & 7.296$\pm$0.078 & 0.332$\pm$0.008 & 6.837$\pm$0.011 & 0.345$\pm$0.002 & 7.293$\pm$0.012 & 0.144$\pm$0.003 \\
& learned completeness & 7.400$\pm$0.102 & 0.323$\pm$0.007 & 6.858$\pm$0.003 & 0.339$\pm$0.003 & 7.305$\pm$0.014 & 0.141$\pm$0.003 \\
& max-category-margin$\star$& 6.837$\pm$0.090 & 0.423$\pm$0.004 & 6.549$\pm$0.016 & 0.413$\pm$0.004 & 6.930$\pm$0.011 & 0.265$\pm$0.000 \\
\midrule
\multirow{2}{*}{Labeling} & component-mean category& 7.667$\pm$0.059 & 0.262$\pm$0.003 & 7.227$\pm$0.029 & 0.246$\pm$0.005 & 7.562$\pm$0.012 & 0.053$\pm$0.004 \\
& representative raw cosines$\star$& 6.837$\pm$0.090 & 0.423$\pm$0.004 & 6.549$\pm$0.016 & 0.413$\pm$0.004 & 6.930$\pm$0.011 & 0.265$\pm$0.000 \\
\midrule
\multirow{3}{*}{Admission} & none & 6.837$\pm$0.090 & 0.423$\pm$0.004 & 6.549$\pm$0.016 & 0.413$\pm$0.004 & 6.930$\pm$0.011 & 0.265$\pm$0.000 \\
& background-margin$\star$& 6.793$\pm$0.051 & 0.426$\pm$0.007 & 6.485$\pm$0.058 & 0.416$\pm$0.005 & 6.715$\pm$0.123 & 0.283$\pm$0.011 \\
& cosine + margin conjunction & 6.785$\pm$0.046 & 0.424$\pm$0.004 & 6.467$\pm$0.014 & 0.412$\pm$0.000 & 6.680$\pm$0.018 & 0.287$\pm$0.002 \\
\bottomrule
\end{tabular*}
\caption{Complete decoder selection and frozen target-set audit. Inner45 is the only split used for selecting graph construction, representative, labeling, and admission variants. In the graph block, the representative is maximum-category-margin, labeling uses its raw cosines, and admission is disabled. The representative block carries forward the starred graph and uses representative raw-cosine labeling without admission; the labeling block additionally carries forward the starred representative; and the admission block carries forward all preceding starred choices. The clean500 and OmniCount-sub columns are reported only after all choices are frozen. Stars denote the variants selected on inner45; target-set winners are not used for model selection.}
\label{tab:decoder_ablation_supp}
\end{table*}

\subsection{Training and Decoding}

The category head remains fixed. Every reliable unordered pair $\{i,j\}$ contributes the directed rows $\boldsymbol\phi_{i\rightarrow j}$ and $\boldsymbol\phi_{j\rightarrow i}$, giving 91,016 rows from 45,508 pairs. Same-category and same-instance targets are duplicated across directions, so their directed positive/negative supports are 74,838/16,178 and 45,628/45,388. Proposal-completeness targets remain directional, with 1,269 positives and 89,747 negatives: $y^{\mathrm{comp}}_{i\rightarrow j}=1$ only when $j$ is the higher-coverage proposal satisfying the containment and coverage-gap rule. For branch $r\in\{\mathrm{sem,inst,comp}\}$, let $\mathbf z^r$ and $\mathbf y^r$ collect the logits and targets of its directed rows, and let $N_r^+$ and $N_r^-$ denote its positive and negative supports. The \texttt{BCEWithLogits} call uses $\mathrm{pos\_weight}=\lambda_r$, where $\lambda_r=N_r^-/N_r^+$. Writing this weighted loss as $\mathrm{BCEWithLogits}_{\lambda_r}$, the training objective is
\begin{equation}
\begin{aligned}
\ell_r&=\mathrm{BCEWithLogits}_{\lambda_r}
(\mathbf z^r,\mathbf y^r),\\
\mathcal L_{\mathrm{rel}}&=
\sum_{r\in\{\mathrm{sem,inst,comp}\}}\ell_r.
\end{aligned}
\end{equation}
Binary cross-entropy is applied to every directional logit before any direction averaging. The relation head is trained for 80 deterministic full-batch epochs with AdamW, learning rate $3\!\times\!10^{-3}$, weight decay $10^{-4}$, and dropout 0.1. At inference, the same-category and same-instance logits are averaged across directions before the sigmoid, whereas completeness retains both directions for its diagnostic representative rule. Decoder choices and thresholds are selected only on the disjoint inner45 split using the staged criteria in the main paper.

UniCounting uses the same-instance score from the jointly trained relation head, top2-OR sparsification, union-find components, maximum-category-margin representative selection, representative raw-cosine labeling, and seven-concept background-margin admission. For relation-head seeds 17, 42, and 73, $\tau_{\rm inst}^{(s)}$ is 0.410, 0.400, and 0.400, and $\tau_{\rm bg}^{(s)}$ is $-0.045626$, $-0.035712$, and $-0.017443$. An admitted component adds one to its assigned category; a rejected component adds zero everywhere. Candidate indices follow fixed cache order, pair indices follow ascending upper-triangular enumeration, and vocabulary indices follow the released class list. Exact ties use the lower pair, candidate, or vocabulary index, respectively; equality at either threshold is accepted, and all eligible edges are retained when an endpoint has fewer than two.

\paragraph{Diagnostic decoder definitions.}
Let $a^{\rm sem}_{ij}$ and $a^{\rm inst}_{ij}$ be the direction-averaged sigmoid outputs. The ``hard semantic $\land$ instance'' graph admits pair $\{i,j\}$ exactly when $a^{\rm sem}_{ij}\geq\tau_{\rm sem}^{(s)}$ and $a^{\rm inst}_{ij}\geq\tau_{\rm inst}^{(s)}$, ranks eligible edges by $\sqrt{a^{\rm sem}_{ij}a^{\rm inst}_{ij}}$, and then applies top2-OR. For seeds 17/42/73, $\tau_{\rm sem}^{(s)}$ is 0.920/0.940/0.940 and $\tau_{\rm inst}^{(s)}$ is 0.410/0.400/0.400. Exact edge-score ties use the lower ascending upper-triangular pair index.

For learned-completeness representative selection, let $p^{\rm comp}_{i\rightarrow j}=\sigma(z^{\rm comp}_{i\rightarrow j})$, where $\sigma$ is the logistic sigmoid. Within a component $G$, each candidate receives
\begin{equation}
R_i=\sum_{j\in G\setminus\{i\}}\left(p^{\rm comp}_{j\rightarrow i}-p^{\rm comp}_{i\rightarrow j}\right),
\end{equation}
and the representative maximizes $R_i$; ties are broken by the candidate's largest raw category cosine and then by the negative candidate index. Thus, exact remaining ties select the lower candidate index.

For component $G$, define the representative-evidence vector $\mathbf s_G=(s_{k_G^\star c})_{c\in\mathcal V}$, and let $s_{G,(1)}$ and $s_{G,(2)}$ be its largest and second-largest entries. The cosine--margin conjunction admits $G$ when $s_{G,(1)}\geq\tau_{\rm cos}^{(s)}$ and $s_{G,(1)}-s_{G,(2)}\geq\tau_{\rm mar}^{(s)}$. For seeds 17/42/73, $(\tau_{\rm cos}^{(s)},\tau_{\rm mar}^{(s)})$ is $(0.243097,0.011030)$, $(0.235304,0.006040)$, and $(0.243339,0.007480)$. For component-support F1, one decoded component is one binary item: its target is positive exactly when it contains at least one reliable proposal, and its prediction is positive exactly when the conjunction admits it. Component-level TP, FP, and FN counts are pooled across inner45. Each threshold pair is selected from the 41-quantile Cartesian grid by this F1, then precision, then the smaller sum of the two thresholds. Across admission rules, mean inner45 micro-F1 is the selection metric and lower vector $\ell_1$ is the tie-break.

\section{Extended Results}

\subsection{Evaluation Splits and Detector Protocols}

Clean500 is formed by sorting the 4,940 eligible val2017 file names by \texttt{SHA256(salt|file\_name)}, after excluding the clean60 diagnostic identities, and taking the first 500. The salt is \texttt{exp-v10-coco-val-clean500-v1-20260725}. The selection procedure reads no annotation, prediction, or metric, and clean500 overlaps clean60 in zero images. All thresholds are selected on inner45 and frozen before preprocessing any clean500 or transfer-set image. Ground-truth annotations are loaded only after the prediction files have been finalized.

Unless a table states otherwise, relation-head rows report the mean and sample standard deviation of the three relation-seed point estimates. Clean500 macro-F1 support contains 78 of the 80 COCO categories; \emph{toaster} and \emph{hair drier} are excluded because they have no positive ground-truth cells. Each paired confidence interval uses 10,000 image-bootstrap replicates with seed 20260725. Within a replicate, the same resampled image multiplicities are applied to every method and relation seed, each metric is recomputed separately for each seed, and the seed-wise metrics are then averaged. Macro-F1 support is recomputed within every replicate, and categories with zero positives in that replicate are excluded from its macro average. The reported 95\% interval is the percentile interval over the seed-averaged replicates.

OWLv2-All80 \citep{minderer2023owlv2} uses the fixed \texttt{google/owlv2-base-patch16-ensemble} checkpoint at the revision whose SHA begins \texttt{cfd3195b}; the full hash is provided with the accompanying artifacts. All 80 raw official COCO thing-category names are passed in one identical list for every image, without aliases or a prompt ensemble. Snapshot-native $960\times960$ preprocessing and processor decoding are followed by classwise \texttt{batched\_nms} at IoU 0.50. Every detection with score strictly above 0.350 contributes one count. The threshold is selected from 201 points on inner45 by minimum vector $\ell_1$, then maximum micro-F1 and the higher threshold as ties. Its inner45 vector $\ell_1$ is 3.6889 and micro-F1 is 0.7843. Clean500 is not used for calibration.

The fully supervised reference uses torchvision Faster R-CNN with a ResNet-50-FPN v2 backbone and the \texttt{FasterRCNN\_ResNet50\_FPN\_V2} \texttt{COCO\_V1} weights \citep{ren2015fasterrcnn}. The released checkpoint SHA-256 begins \texttt{dd69338a}. Official preprocessing is followed by explicit classwise NMS at IoU 0.50, and detections are mapped from the torchvision label space to the canonical COCO80 coordinates. A strict score threshold of 0.775 is selected from 191 points on inner45 by minimum complete-vector $\ell_1$, then maximum micro-F1 and the higher threshold as ties. Its inner45 vector $\ell_1$ is 2.3111 and micro-F1 is 0.8800. No clean500 annotation or metric is accessed before its predictions are sealed. This model uses full COCO box supervision and is therefore a supervised reference rather than a matched-training baseline.

\subsection{Controlled Edge and Grouping Ablation}
\label{supp:edge_ablation}

The controlled study fixes the candidate pool and unordered-pair universe together with the top2-OR rule, connected-component construction, maximum-category-margin representative rule, representative raw-cosine labeling rule, and run-matched background-margin admission rule. Five score rows apply these procedures after replacing only the pair score and its scalar threshold. The singleton control removes every edge, while the overlap guard additionally requires strictly positive mask intersection. A single threshold shared across seeds 17, 42, and 73 is selected for each score on the same 45-image inner-validation split. Selection minimizes three-seed mean vector $\ell_1$, with mean micro presence F1 and then the higher threshold used as ties. The target tests are not consulted.

Table~2 of the main paper reports end-task metrics for the five edge scores together with the singleton and overlap-guard controls, while Table~S8 reports their full partition diagnostics. Mask IoU uses the common threshold 0.005 selected in Table~S4. The relation-head score uses the common threshold $\tau=0.465$ selected in the same table, whereas UniCounting uses the run-specific thresholds listed above.

\paragraph{Grouping diagnostic protocol.}
All grouping metrics are computed before category admission. Ground-truth masks are used only for post hoc evaluation, not for prediction. Let $\mathcal R$ denote candidates satisfying the reliable-match rule above, retain $u_i$ for the matched non-ignore GT instance of reliable candidate $i$, and let $\mathcal C$ be the component partition. For $C\in\mathcal C$, define $\mathcal R_C=C\cap\mathcal R$ and $n_{Cu}=|\{i\in\mathcal R_C:u_i=u\}|$. Only components in $\mathcal C_R=\{C:|\mathcal R_C|>0\}$ enter purity and merge error.

For Pair F1, every unordered pair of reliable candidates is GT-positive when its two matched-instance indices agree and predicted-positive when both candidates lie in the same component. Counts are pooled across all images before computing
\begin{equation}
\begin{aligned}
P_{\rm pair}&=\frac{TP}{TP+FP},\\
R_{\rm pair}&=\frac{TP}{TP+FN},\\
F_{\rm pair}&=\frac{2TP}{2TP+FP+FN}.
\end{aligned}
\end{equation}
Let $d_C=|\{u:n_{Cu}>0\}|$. Component purity and merge error are
\begin{equation}
\begin{aligned}
\mathrm{Purity}&=\frac{\sum_{C\in\mathcal C_R}\max_u n_{Cu}}{\sum_{C\in\mathcal C_R}|\mathcal R_C|},\\
\mathrm{Merge}&=\frac{\sum_{C\in\mathcal C_R}\mathbf 1[d_C>1]}{|\mathcal C_R|}.
\end{aligned}
\end{equation}
Thus, purity is weighted by reliable-candidate count, whereas merge error gives each evaluable component one vote. Let $\mathcal U=\{u_i:i\in\mathcal R\}$ be the supported GT instances and $k_u=|\{C:n_{Cu}>0\}|$. Components per instance and Fragmented are
\begin{equation}
\begin{aligned}
\mathrm{Comp/inst.}&=\frac{\sum_{u\in\mathcal U}k_u}{|\mathcal U|},\\
\mathrm{Fragmented}&=\frac{\sum_{u\in\mathcal U}\mathbf 1[k_u>1]}{|\mathcal U|}.
\end{aligned}
\end{equation}
Their denominator excludes annotated instances with no reliable candidate. For every component, define
\begin{equation}
\begin{aligned}
u_C^\star&=\arg\max_{u\ \mathrm{nonignore}}
\sum_{i\in C}|m_i\cap G_u|,\\
\mathcal C_{\rm IoU}&=\left\{C:
\sum_{i\in C}|m_i\cap G_{u_C^\star}|>0\right\},
\end{aligned}
\end{equation}
with the lower stored GT index breaking exact ties. If $r(C)$ is the selected representative, the reported aggregate is
\begin{equation}
\mathrm{RepIoU}
=\frac{1}{|\mathcal C_{\rm IoU}|}
\sum_{C\in\mathcal C_{\rm IoU}}
\operatorname{IoU}(m_{r(C)},G_{u_C^\star}).
\end{equation}
Thus, each evaluable component across all images receives one vote. Pair counts are pooled before F1, and each reported relation-head metric is computed per seed and then averaged over the three relation-head seeds.

Candidate recall uses all retained pre-grouping proposals, without the reliable-candidate filter. For non-ignore ground-truth instance $G_u$, let $b_u=\max_i\operatorname{IoU}(m_i,G_u)$ over every candidate $m_i$ from the same image. At threshold $t$, the reported micro recall is
\begin{equation}
R_{\mathrm{cand}}(t)=
\frac{\sum_{u\in\mathcal G}\mathbf 1[b_u\geq t]}{|\mathcal G|},
\end{equation}
where $\mathcal G$ pools all non-ignore instances across clean500. Thus 62.4\% at $t=0.50$ and 48.8\% at $t=0.75$ measure the candidate-generation ceiling before edge scoring, grouping, representative selection, labeling, or admission.

\subsection{Complete-Vector Transfer Diagnostic}

OmniCount-sub is a sparse transfer diagnostic in which the zero vector obtains lower vector $\ell_1$ by predicting no positive coordinates, while UniCounting recovers nonzero category presence. This contrast shows why vector error should be interpreted jointly with presence F1.

\nobibliography{aaai2027}